\documentclass[sigconf,nonacm]{acmart}
\AtBeginDocument{%
  }

\setcopyright{none}
\renewcommand\footnotetextcopyrightpermission[1]{}

\usepackage{url}            
\usepackage{booktabs}       

\usepackage{amsmath,amssymb} 
\usepackage{nicefrac}       
\usepackage{microtype}      
\usepackage{xcolor}         
\usepackage{multirow}
\usepackage{enumitem, tabularx}
\usepackage{wrapfig,lipsum,booktabs}
\usepackage{bbding}
\usepackage{enumitem, tabularx}
\usepackage{adjustbox}
\usepackage{tcolorbox}
\usepackage{listings}
\usepackage{booktabs}
\usepackage{colortbl}
\definecolor{TableHeader}{RGB}{239,242,245}
\definecolor{TableBest}{RGB}{242,242,242}
\newcolumntype{Y}{>{\raggedright\arraybackslash}X}

\begin{document}

\title{Explainable Cross-Disease Reasoning for Cardiovascular Risk Assessment from Low-Dose Computed Tomography}

\author{Yifei Zhang}
\email{yifei.zhang2@emory.edu}
\affiliation{
  \institution{Department of Computer Science, Emory University}
  \city{Atlanta}
  \state{GA}
  \country{USA}
}

\author{Jiashuo Zhang}
\email{jzhan427@jhu.edu}
\affiliation{
  \institution{Department of Computer Science, Johns Hopkins University}
  \city{Baltimore}
  \state{MD}
  \country{USA}
}

\author{Mojtaba Safari}
\email{mojtaba.safari@emory.edu}
\affiliation{
  \institution{Department of Radiation Oncology and Winship Cancer Institute, Emory University}
  \city{Atlanta}
  \state{GA}
  \country{USA}
}

\author{Xiaofeng Yang}
\email{xiaofeng.yang@emory.edu}
\affiliation{
  \institution{Department of Radiation Oncology and Winship Cancer Institute, Emory University}
  \city{Atlanta}
  \state{GA}
  \country{USA}
}

\author{Liang Zhao}
\email{liang.zhao@emory.edu}
\affiliation{
  \institution{Department of Computer Science, Emory University}
  \city{Atlanta}
  \state{GA}
  \country{USA}
}

\renewcommand{\shortauthors}{Zhang et al.}

\begin{abstract}
Low-dose chest computed tomography (LDCT) captures pulmonary and cardiac structures in a single scan, enabling joint assessment of lung and cardiovascular health. Existing approaches typically model these domains independently and do not explicitly represent their physiological interactions. We propose an Explainable Cross-Disease Reasoning Framework for cardiovascular risk assessment from LDCT. The framework follows a constrained clinical-information pathway: it extracts pulmonary findings, grounds cross-organ mechanisms in medical knowledge, and produces a cardiovascular prediction with a natural-language rationale. It combines four components: a frozen lung-risk prior, a pulmonary perception module, an agentic reasoning module, and a cardiac subvolume feature extractor. Their outputs are fused to integrate localized cardiac evidence with mechanism-level pulmonary context. On the National Lung Screening Trial cohort, the framework achieves an AUC of 0.919 for CVD screening and up to 0.838 for CVD mortality prediction, outperforming cardiac-specific, single-disease, and foundation-model baselines. Targeted controls indicate that the gains are not explained by additional thoracic visual features alone, fixed rule propagation, or a single reasoning backend. The proposed framework thus provides an auditable approach to cross-disease cardiovascular risk assessment from LDCT.
\end{abstract}




\maketitle

\section{Introduction}
\label{sec:intro}

Lung cancer and cardiovascular disease (CVD) are the two leading causes of death worldwide, jointly accounting for over one-third of annual mortality~\cite{naghavi2024global}. In high-risk populations such as long-term smokers, these diseases exhibit complex pathophysiological interplay beyond shared lifestyle factors. Epidemiological evidence supports a bidirectional link: lung cancer is an independent risk factor for coronary artery disease (CAD), while patients with heart failure have a significantly increased risk of incident lung cancer (HR 1.89)~\cite{de2024cardiovascular}. This intersection is especially striking in screening cohorts; in the National Lung Screening Trial (NLST), cardiovascular deaths (486) outnumbered lung cancer deaths (356) during follow-up~\cite{chao2021deep, chiles2015association}. Although both share smoking as a dominant risk factor, their exposure--response patterns differ: lung cancer risk increases nearly linearly with cumulative exposure, whereas CVD mortality rises steeply at low exposure and then plateaus~\cite{pope2011lung}. These trends reflect distinct yet interconnected mechanisms, including chronic systemic inflammation and oxidative stress, leaving cancer patients with up to a ten-fold higher risk of CVD mortality than the general population~\cite{chao2021deep, de2024cardiovascular}.

\begin{figure}[!ht]
\centerline{\includegraphics[width=\columnwidth]{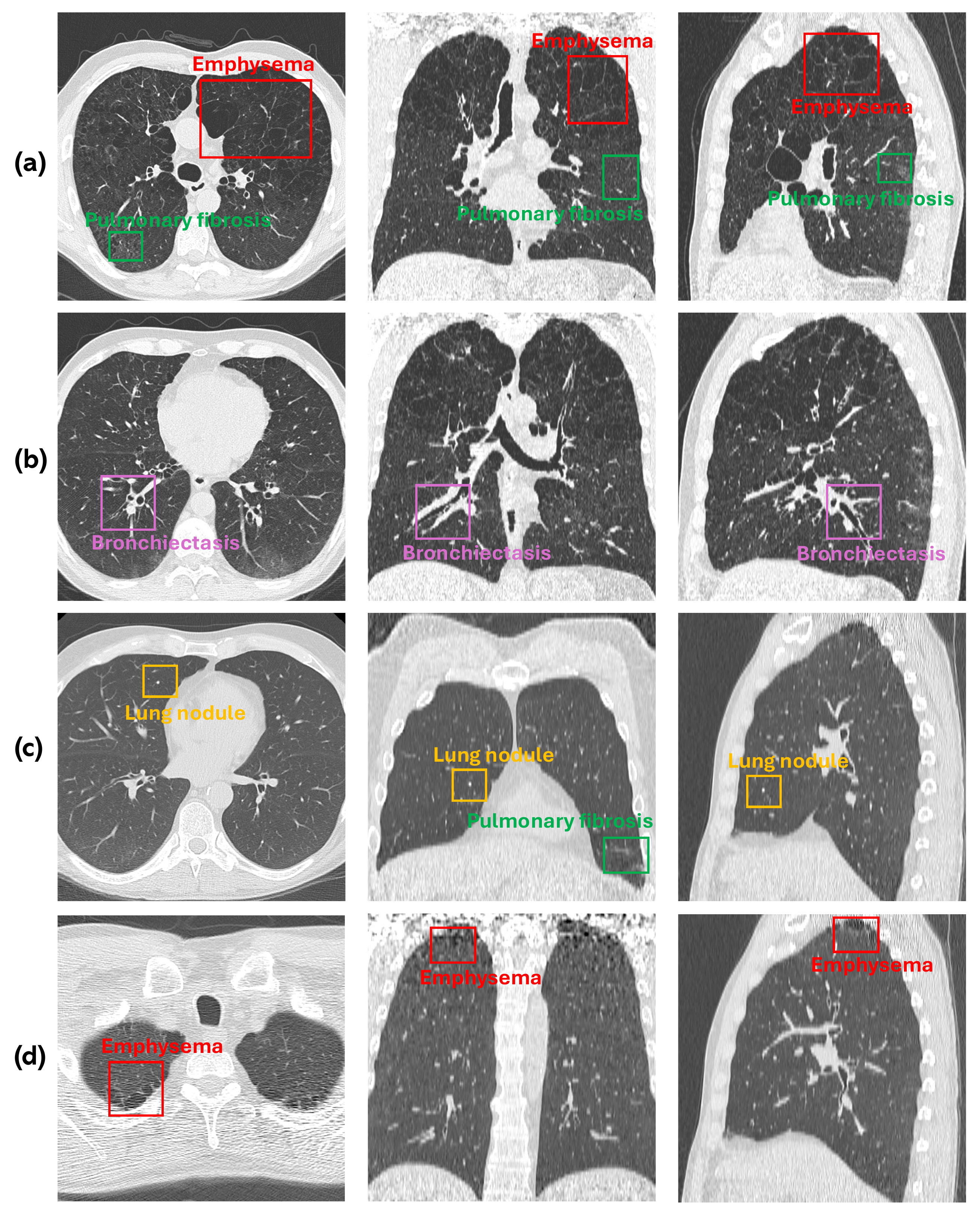}}
\caption{\mdseries \textbf{Low-Dose Chest CT as a Shared Window for Cardiopulmonary Assessment.} A single LDCT acquisition captures both pulmonary integrity and cardiac anatomy within one volumetric scan. Panels (a--d) show representative pulmonary abnormalities across distinct lung regions in axial, coronal, and sagittal planes. These examples highlight the spatial coexistence of heterogeneous lung pathologies and the cardiac silhouette, motivating our unified cardiopulmonary risk prediction framework.}
\label{fig:teaser}
\end{figure}

Low-dose chest computed tomography (LDCT) is the standard of care for lung cancer screening, with the NLST showing a 20\% reduction in lung cancer-specific mortality. Subsequent trials, including NELSON and UKLS, have consistently validated this survival benefit~\cite{national2011reduced, de2020reduced, field2021lung}. Beyond its primary oncologic utility, LDCT offers an underexploited window into the cardiopulmonary nexus. Clinical reporting guidance already recommends reviewing cardiovascular and other clinically significant non-lung findings on thoracic or lung-screening CT, including coronary and cardiac calcification~\cite{williams2021reporting,tanoue2022standardizing}. Recent data indicate that up to 60.1\% of lung cancer screening candidates already harbor coronary artery calcification (CAC)~\cite{park2024coronary}; real-world reporting of such findings can affect downstream prevention, including statin prescription~\cite{hammer2023reporting}. However, the prognostic value of LDCT extends beyond cardiac regions alone. Emerging evidence suggests that specific pulmonary abnormalities are intrinsically linked to cardiovascular health. For instance, quantitative emphysema and interstitial lung abnormalities (ILAs) detected on LDCT are robust, independent predictors of CVD mortality, even after adjustment for traditional risk factors such as CAC~\cite{balbi2023automated, de2024cardiovascular}; recent long-term evidence further shows that baseline emphysema on LDCT predicts cardiovascular mortality over extended follow-up~\cite{gonzalez2025emphysema}. This suggests that lung parenchymal morphology contains critical prognostic cues for the heart, reflecting shared mechanisms driven by systemic inflammation and endothelial dysfunction. Thus, a single LDCT scan can capture a combined cardiopulmonary signature. Fig.~\ref{fig:teaser} illustrates this concept by highlighting concurrent pulmonary abnormalities and cardiac risk markers within the same 3D volumetric scan.

Despite this pathophysiological crosstalk, current computational research has not fully leveraged this medical insight. The main bottleneck is the isolated modeling of interconnected systems: predominant approaches treat lung cancer detection and CVD risk stratification as separate tasks, analyzing pulmonary nodules and cardiac structures independently. By overlooking pulmonary biomarkers for cardiac risk, these methods discard critical cross-organ information. Consequently, three major challenges hinder effective dual screening from LDCT. First, CVD mortality discrimination from cardiac regions alone remains moderate~\cite{chiles2015association,sabia2023fully}, indicating that organ-restricted features fail to capture the broader systemic disease context. Second, cross-organ reasoning is not simple feature concatenation: the relation between pulmonary findings and cardiovascular outcomes is latent, multi-step, and only weakly supervised by outcome labels. A useful model must transform patient-specific pulmonary evidence into clinically grounded cardiovascular hypotheses rather than merely aggregate embeddings. Third, although recent agentic reasoning approaches introduce clinically inspired inference mechanisms~\cite{li2024agent,sun2025cpathagent,zheng2024well}, they remain limited to single-organ analysis and rarely assess whether generated rationales are useful beyond plausible text. Together, these limitations constrain predictive accuracy and clinical utility, motivating a unified, interpretable approach to joint cardiopulmonary analysis.

To address these challenges and bridge radiological isolation with physiological interdependence, we propose an \emph{Explainable Cross-Disease Reasoning Framework} that treats a single LDCT scan as a knowledge-bearing source for cardiopulmonary risk assessment. Our architecture couples four clinically grounded components: a \textit{Lung-Risk Prior} for systemic pulmonary risk context, a \textit{Pulmonary Perception Module} for structured lung abnormality findings, an \textit{Agentic Pulmonary-to-Cardiac Reasoning Module} for literature-grounded mechanism synthesis, and a \textit{Cardiac Localization and Feature Extraction Module} for local cardiac morphology. A final \textit{Multimodal Fusion Module} integrates these visual and reasoning-derived signals. This design shifts the LLM from a generic text aggregator to a constrained knowledge-reasoning component for latent inter-organ relationships.

Our contributions are threefold:
\begin{itemize}[leftmargin=1.25em,itemsep=0.12em,topsep=0.25em,parsep=0pt,partopsep=0pt]
    \item We formulate LDCT-based cardiovascular risk prediction as an explicit cross-disease reasoning problem, converting pulmonary findings into mechanism-level cardiovascular hypotheses rather than treating them as unstructured auxiliary features.
    \item We develop a knowledge-guided multimodal framework that anchors final prediction in localized cardiac visual evidence while exposing an auditable pulmonary-to-cardiac reasoning pathway.
    \item We provide a targeted evaluation suite for this setting, including cardiac-specific baselines, vision-only fusion controls, structured semantic controls, fixed-rule propagation, and reasoning-backend ablations.
\end{itemize}
Extensive experiments on the NLST cohort show state-of-the-art performance for CVD screening (AUC 0.919) and mortality prediction (up to AUC 0.838). Targeted controls further show that the gain is not explained by adding lung visual embeddings alone, fixed rule propagation, or a single proprietary reasoning backend. Qualitative analyses via Grad-CAM and textual attention provide complementary evidence that learned decisions align with interpretable cardiac regions and mechanism-level pulmonary-to-cardiac rationales.

\section{Related Work}
\label{sec:related}

\subsection{Lung Cancer Risk Prediction from LDCT}

Low-dose chest CT is the most effective modality for lung cancer screening, with large randomized trials such as NLST, NELSON, and UKLS demonstrating significant reductions in lung cancer mortality~\cite{national2011reduced,de2020reduced,field2021lung}. These successes have motivated extensive research on automated lung cancer risk prediction from LDCT~\cite{veasey2025low}. Early approaches typically followed a two-stage approach, combining pulmonary nodule detection with malignancy risk estimation based on handcrafted or learned features~\cite{setio2016pulmonary,mcwilliams2013probability}. With the advent of deep learning, convolutional models trained on large-scale datasets such as LIDC-IDRI and LUNA16 achieved strong performance for nodule detection and characterization, with 3D architectures generally outperforming their 2D counterparts~\cite{armato2011lung,setio2017validation,jiang2025benchmark}. More recent work has shifted toward end-to-end longitudinal risk estimation. Notably, \textit{Sybil} directly predicts 1--6 year lung cancer incidence from volumetric LDCT and demonstrates robust generalization across external cohorts~\cite{mikhael2023sybil}. Related systems have also been validated for incidental nodule detection in non-screening CT populations, highlighting their applicability beyond controlled trial settings~\cite{hendrix2023deep}. Despite these advances, most lung cancer risk models remain narrowly optimized for oncologic outcomes, leaving non-oncologic information embedded in LDCT, particularly signals relevant to cardiovascular health, largely underutilized.

\subsection{Cardiovascular Risk Prediction from LDCT}

Beyond pulmonary structures, LDCT inherently captures the heart and thoracic vasculature, enabling opportunistic cardiovascular risk prediction. Early studies focused on quantifying coronary artery calcium (CAC), thoracic aortic calcification, and pericardial fat, which are well-established biomarkers of cardiovascular disease (CVD) risk~\cite{chiles2015association,pope2011lung,foldyna2024deep}. In lung cancer screening cohorts, CAC burden has been shown to strongly predict all-cause and cardiovascular mortality, often rivaling lung cancer as a cause of death~\cite{jacobs2012coronary,de2020reduced}. Building on this foundation, recent deep learning methods have automated calcium scoring and extended LDCT analysis toward direct outcome prediction. Representative examples include DeepCAC~\cite{zeleznik2021deep}, KAMP-Net~\cite{guo2019knowledge}, and related frameworks for fully automated CAC quantification~\cite{lessmann2017automatic}, which achieve moderate discrimination for cardiovascular mortality in the NLST. More direct mortality-oriented studies have used heart-region autoencoder features to predict 5-year cardiovascular mortality~\cite{van2019direct}, automated CT-derived risk factors such as CAC, liver steatosis, and emphysema for CVD risk assessment~\cite{stemmer2020using}, and 3D ResNet models for long-term non-lung-cancer mortality prediction in NLST~\cite{lu2022deep}. Beyond coronary calcification, multiple screening cohorts have shown that emphysema extent, muscle mass, and fat attenuation provide complementary prognostic signals as systemic markers of cardiometabolic health~\cite{sabia2023fully,balbi2023automated,mascalchi2023pulmonary}. Nevertheless, most existing approaches emphasize cardiac structures, automated risk-factor quantification, or opaque image-level mortality prediction, and do not explicitly integrate pulmonary abnormalities that co-vary with cardiovascular outcomes, limiting their ability to capture the multifactorial disease risk encoded in LDCT.

\subsection{Cross-Disease Learning from LDCT}

Although lung cancer and cardiovascular disease (CVD) are often studied in isolation, their shared risk factors and frequent co-occurrence motivate cross-disease learning from LDCT. Early work, such as Tri2D-Net, demonstrated that LDCT encodes both pulmonary and cardiac signals by jointly leveraging coronary calcium and pericardial fat to predict cardiovascular mortality in the NLST, with performance comparable to dedicated calcium scoring~\cite{chao2021deep}, establishing the feasibility of cross-disease modeling from screening scans. More recently, large-scale foundation models have extended this framework toward multitask learning across heterogeneous clinical endpoints. M3FM, trained on over 120{,}000 CT scans with paired reports, achieved strong performance on lung cancer risk prediction, coronary artery disease classification, and CVD mortality~\cite{niu2025medical}. Merlin further integrated volumetric CT with EHR codes and reports to support a broad range of downstream clinical tasks~\cite{blankemeier2024merlin}. In parallel, Hamamci \textit{et al.} introduced a generalist 3D CT foundation model aligning CT images with radiology text via CT-CLIP and CT-CHAT, enabling zero-shot pulmonary finding recognition and instruction-driven report generation~\cite{hamamci2024developing}. Prompt-based adaptation has also enhanced biomedical foundation models for segmentation~\cite{li2024multi}, although segmentation does not itself represent cross-organ outcome reasoning. Related vision--language models such as MedCLIP~\cite{wang2022medclip}, BioMedCLIP~\cite{zhang2023biomedclip}, RadFM~\cite{wu2025towards}, and CheXagent~\cite{chen2024chexagent} reflect the broader trend toward multimodal pretraining in radiology, though most focus on 2D chest radiographs. Methods such as ESSA further improve explanation-aware biomedical prediction through saliency-guided supervision~\cite{gu2023essa}, but do not expose a structured pulmonary-to-cardiac mechanism interface. Despite this progress, existing cross-disease and multitask models largely function as black boxes, offering only coarse post-hoc attribution. In contrast, our framework explicitly models pulmonary-to-cardiac reasoning through interpretable intermediate indicators, enabling transparent cross-disease inference for cardiovascular risk assessment from LDCT.

\begin{figure*}[!ht]
\centerline{\includegraphics[width=0.98\linewidth]{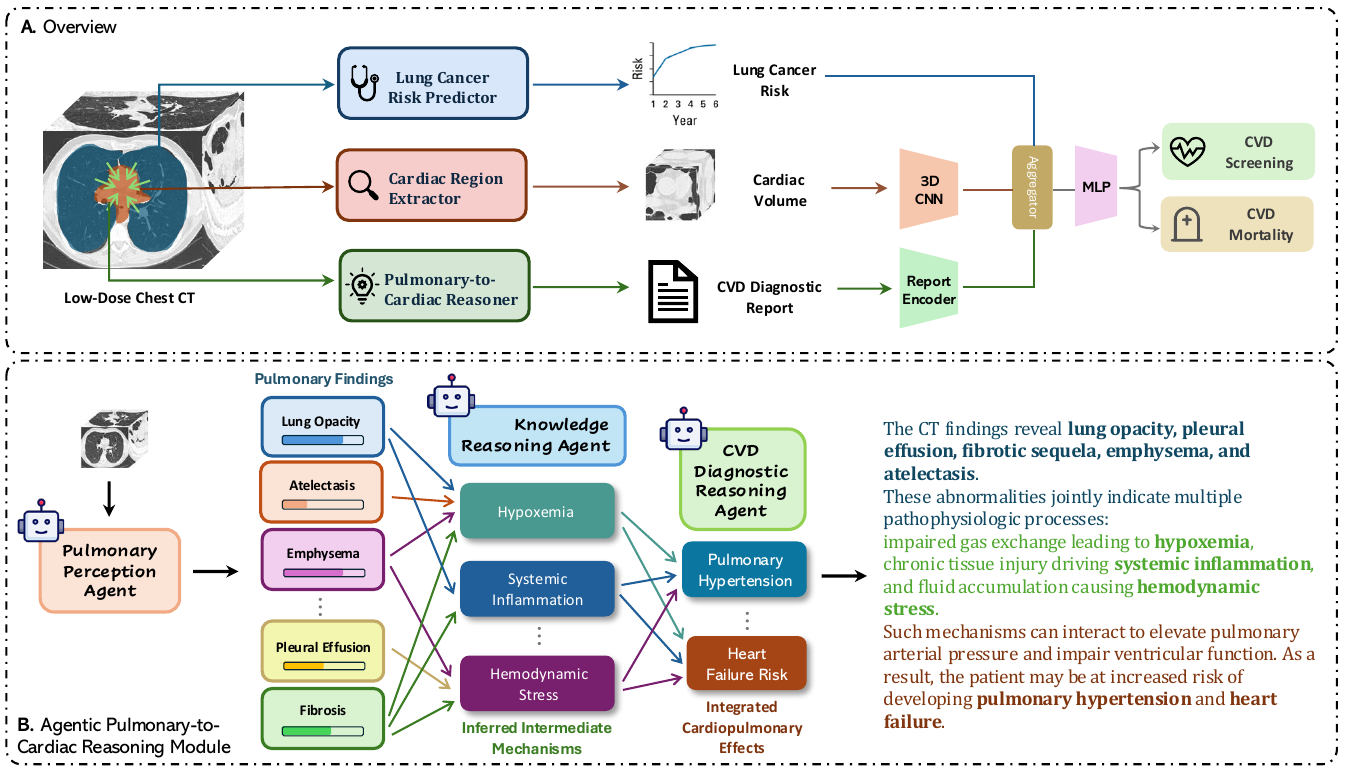}}
\caption{\mdseries
\textbf{Overview of the proposed explainable cross-disease reasoning framework.} \textbf{A.} Four modules are organized into three pathways: lung-risk prior, cardiac feature extraction, and pulmonary perception-to-reasoning. Their risk score, cardiac representation, and report embedding are fused for CVD screening and mortality prediction. \textbf{B.} The reasoning module translates pulmonary findings into mechanisms, cardiopulmonary effects, and a concise diagnostic rationale.}
\label{fig:overview}
\end{figure*}

\section{Methodology}
\label{sec:methods}

In this section, we present the details of the proposed \emph{Explainable Cross-Disease Reasoning Framework}. We first formalize the overall pipeline in \S\ref{sec:overview}, followed by the agentic pulmonary-to-cardiac reasoning interface in \S\ref{subsec:pulmonary_to_cardiac_reasoning}. We then describe the visual perception modules for systemic lung risk context and localized cardiac feature extraction in \S\ref{subsec:lung_cancer_risk_analysis} and \S\ref{subsec:cardiac_extraction}, and finally define the fusion objective in \S\ref{subsec:fusion_objective}.

\subsection{Overview of the Explainable Cross-Disease Reasoning Framework}
\label{sec:overview}

We formulate LDCT-based CVD assessment as a supervised multimodal prediction problem over a subject-level cohort
\[
\mathcal{D}=\{(X_i,y_i^{\mathrm{scr}},y_i^{\mathrm{mort}})\}_{i=1}^{N},
\]
where $X_i\in\mathbb{R}^{H\times W\times D}$ is an LDCT volume and $y_i^{\mathrm{scr}},y_i^{\mathrm{mort}}\in\{0,1\}$ denote CVD screening and CVD mortality labels. Because cardiovascular risk is not fully localized to a manually defined cardiac region, our four components decompose the mapping $X_i\mapsto \hat{y}_i$ into three fused representations:
\[
\begin{aligned}
z_i^{\mathrm{lung}} &= \mathcal{F}_{\mathrm{lung}}(X_i),\\
z_i^{\mathrm{card}} &= \mathcal{E}_{\mathrm{card}}(\mathcal{D}_{\mathrm{card}}(X_i);\theta_c),\\
z_i^{\mathrm{reason}} &= \mathcal{E}_{\mathrm{text}}(\mathcal{A}_{\mathrm{cross}}(R_i;\mathcal{C})),
\end{aligned}
\]
where $z_i^{\mathrm{lung}}$ is a frozen pulmonary risk prior, $z_i^{\mathrm{card}}$ is a trainable cardiac-region representation, and $z_i^{\mathrm{reason}}$ encodes a mechanism-level rationale generated from pulmonary findings $R_i$ and corpus $\mathcal{C}$. Thus, pulmonary perception is a separate auditable interface for the reasoning pathway, while final fusion still operates on three representations. As illustrated in Fig.~\ref{fig:overview}, the four components are organized into three interconnected stages:

\paragraph{1. Dual-Pathway Visual Perception}
To capture the heterogeneous nature of cardiopulmonary signals, we employ two specialized visual streams. 
First, a frozen \textbf{Lung Cancer Risk Prior} (\S\ref{subsec:lung_cancer_risk_analysis}) processes the full lung volume to extract a temporal malignancy risk vector $z_{\text{lung}}$, serving as a high-level proxy for chronic systemic inflammation and disease burden. 
Simultaneously, a \textbf{Cardiac Feature Extractor} (\S\ref{subsec:cardiac_extraction}) localizes and processes the cardiac subvolume $X_{\text{card}}$ to encode fine-grained morphological biomarkers (e.g., calcification, chamber geometry) into a dense representation $z_{\text{card}}$. 
This dual-pathway design ensures that both the \textit{systemic context} (from the lung) and \textit{local pathology} (from the heart) are explicitly modeled.

\paragraph{2. Agentic Cross-Disease Reasoning}
To bridge the semantic gap between visual features and biological mechanisms, we employ an \textbf{Agentic Pulmonary-to-Cardiac Reasoning} module (\S\ref{subsec:pulmonary_to_cardiac_reasoning}). 
Unlike models that only provide post-hoc attribution, this module decomposes cross-disease inference into a traceable reasoning path.
Given a structured set of pulmonary findings $R_{\text{lung}}$, the agent leverages a knowledge-grounded Large Language Model (LLM) to infer the pathophysiological pathway:
\[
\mathcal{T}_{\text{reason}}: R_{\text{lung}} \xrightarrow{\text{Medical Knowledge}} \text{Mechanisms} \xrightarrow{\text{Synthesis}} E_{\text{cvd}}.
\]
This generates a natural-language rationale $E_{\text{cvd}}$ that articulates how specific lung abnormalities (e.g., fibrosis) may indicate cardiac stress pathways (e.g., hemodynamic overload), aligning the model's intermediate logic with established medical literature without treating the generated text as causal proof.

\paragraph{3. Knowledge-Aware Multimodal Fusion}
The final stage synthesizes these complementary information streams via the Multimodal Cardiovascular Risk Predictor. We employ a semantic encoder to project the textual reasoning $E_{\text{cvd}}$ into a latent embedding $z_{\text{reason}}$, aligning linguistic medical insights with the visual representations $z_{\text{lung}}$ and $z_{\text{card}}$. 
By concatenating normalized vectors into a unified latent space $z_{\text{fusion}}$, the model learns a joint decision boundary that is guided by both imaging textures and reasoning-derived logic:
\[
\begin{aligned}
z_{\text{fusion}} &=
\operatorname{LN}(z_{\text{card}}) \oplus
\operatorname{LN}(z_{\text{lung}}) \oplus
\operatorname{LN}(z_{\text{reason}}),\\
\hat{y}_{\text{cvd}} &= \mathcal{G}_{\text{fusion}}(z_{\text{fusion}}).
\end{aligned}
\]
The trainable visual and fusion components are optimized under the same subject-level splits, while frozen perception and language encoders provide stable pulmonary and textual representations.

\vspace{1ex}
\noindent
In summary, our framework establishes a structured \textit{reasoning-aware pathway} where pulmonary observations contextualize cardiac signals, yielding a prediction system that is accurate, auditable, and aligned with knowledge-intensive multimodal learning.

\subsection{Agentic Pulmonary-to-Cardiac Reasoning}
\label{subsec:pulmonary_to_cardiac_reasoning}

A central component of our framework is the \emph{Agentic Pulmonary-to-Cardiac Reasoning} module. We implement this module as a typed reasoning interface rather than an unconstrained report generator. Given a pulmonary finding set $R_{\text{lung}}$, a retrieval corpus $\mathcal{C}$, and a fixed schema $\Omega$, the module produces a structured cardiovascular reasoning state
\[
\mathcal{A}_{\text{cross}}:(R_{\text{lung}},\mathcal{C},\Omega)
\mapsto \mathcal{S}_{\text{cvd}},
\]
where $\mathcal{S}_{\text{cvd}}$ contains intermediate mechanisms, cardiopulmonary effects, and a bounded rationale. As detailed in Fig.~\ref{fig:overview}B, the interface combines a deterministic pulmonary perception module with two cooperative reasoning agents under fixed input--output contracts.

\textbf{1) Pulmonary Perception Module.}
The perception module imposes a discrete bottleneck between the image encoder and the reasoning module. It translates the LDCT volume $X$ into a set of normalized pulmonary findings:
\begin{equation*}
R_{\text{lung}} = \{(c_j, r_j, s_j) \mid s_j > \tau \}_{j=1}^{N_f},
\end{equation*}
where $c_j$ is a standardized concept, $r_j$ is its textual description (e.g., \textit{``severe centrilobular emphysema''}), and $s_j$ is the detection confidence. We instantiate this stage with CT-CLIP because it was originally benchmarked for zero-shot multi-abnormality detection on internal and external chest CT validation sets, outperforming a supervised CT-Net baseline across key metrics~\cite{hamamci2024developing}. The threshold $\tau=0.5$ is fixed across all experiments and is not tuned on test outcomes. This bottleneck is intentionally restrictive: downstream agents condition only on $R_{\text{lung}}$ and cannot access raw CT voxels, patient identifiers, clinical metadata, or CVD labels. Thus, pulmonary perception errors remain inspectable upstream failures rather than hidden latent states inside the final classifier.

\textbf{2) Knowledge Reasoning Agent.}
The second agent converts pulmonary findings into candidate mechanisms under retrieval grounding. Let $\mathcal{C}=\{d_m\}_{m=1}^{M}$ denote a curated cardiopulmonary knowledge corpus and let $q(R_{\text{lung}})$ be a query formed only from normalized finding terms and mechanism keywords. The agent retrieves the top-$K$ evidence snippets
\[
\mathcal{K}_{\text{lung}} =
\operatorname{TopK}_{d_m\in\mathcal{C}}\,
\operatorname{sim}\!\left(\phi(q(R_{\text{lung}})), \phi(d_m)\right),
\]
where $\phi(\cdot)$ is a text encoder used only for retrieval. In our implementation, $\mathcal{C}$ is fixed before CVD outcome training and consists of citation-backed cardiopulmonary mechanism statements drawn from clinical guidelines, review articles, and radiology literature; it contains no NLST labels, subject identifiers, model predictions, or cohort-level outcome statistics. The agent then constructs a mechanism graph $G_i=(V_i,E_i)$ whose nodes contain pulmonary findings, intermediate mechanisms, and cardiopulmonary effects. Each mechanism edge must be supported by at least one finding--evidence pair:
\[
\forall e\in E_i,\quad
\operatorname{support}(e)
\subseteq R_{\text{lung}}\times \mathcal{K}_{\text{lung}},\quad
|\operatorname{support}(e)|\ge 1.
\]
For example, fibrosis can be linked through retrieved evidence to chronic hypoxemia and then to pulmonary hypertension. This support constraint reduces reliance on the LLM's internal priors and makes unsupported mechanism proposals detectable during audit.

\textbf{3) CVD Diagnostic Reasoning Agent.}
The final agent summarizes the retrieved mechanism graph into a compact reasoning state. Rather than asking the LLM for an unconstrained diagnosis or a standalone CVD label, we define the output as
\[
\mathcal{S}_{\text{cvd}} =
\mathcal{A}_{\text{diag}}(R_{\text{lung}},\mathcal{K}_{\text{lung}},G_i;\Omega)
= (M_{\text{path}}, H_{\text{cvd}}, E_{\text{cvd}}),
\]
where $M_{\text{path}}$ is a set of intermediate mechanisms, $H_{\text{cvd}}$ is a compact set of cardiopulmonary effects, and $E_{\text{cvd}}$ is a bounded natural-language rationale. The schema $\Omega$ specifies required fields, prohibits unsupported patient history, and separates observed findings from inferred mechanisms. Before text encoding, $\mathcal{S}_{\text{cvd}}$ is canonicalized with fixed field order and normalized mechanism names, while explicit screening or mortality decisions are removed from $E_{\text{cvd}}$. The final classifier therefore consumes mechanism text rather than an agent verdict, and all reasoning backends share the same feature schema and downstream supervision.
Formally, the entire reasoning pipeline realizes the mapping:
\[
\begin{aligned}
\mathcal{T}_{\text{cross}}:\quad
X &\xrightarrow{\text{Perception}} R_{\text{lung}}
\xrightarrow{\text{Retrieval}} \mathcal{K}_{\text{lung}}\\
&\xrightarrow{\text{Graph Synthesis}} G_i
\xrightarrow{\text{Schema}} \mathcal{S}_{\text{cvd}}.
\end{aligned}
\]
The example in Fig.~\ref{fig:overview}B illustrates this process: opacity, emphysema, pleural effusion, and fibrosis are mapped to mechanisms such as hypoxemia, systemic inflammation, and hemodynamic stress, and then summarized as cardiopulmonary effects such as pulmonary hypertension and heart failure risk.

\begin{tcolorbox}[
colback=black!1!white,
colframe=black!45!white,
title={Algorithm 1: Cross-Disease Reasoning and Prediction},
fonttitle=\bfseries,
boxrule=0.35pt,
arc=1pt,
left=3pt,
right=3pt,
top=3pt,
bottom=3pt]
\footnotesize
\textbf{Input:} LDCT volume \(X\), cardiopulmonary corpus \(\mathcal{C}\), threshold \(\tau\), trained visual encoders.
\textbf{Output:} CVD risk score \(\hat{y}_{\text{cvd}}\) and rationale \(E_{\text{cvd}}\).
\begin{enumerate}[leftmargin=1.15em,itemsep=0pt,topsep=1pt]
\item Extract pulmonary findings \(R_{\text{lung}}=\{(c_j,r_j,s_j)\mid s_j>\tau\}\).
\item Retrieve evidence \(\mathcal{K}_{\text{lung}}\) from \(\mathcal{C}\) using normalized finding terms.
\item Construct mechanism graph \(G_i\) and bounded reasoning state \(\mathcal{S}_{\text{cvd}}\).
\item Localize cardiac ROI \(X_{\text{card}}\) and encode \(z_{\text{card}}\); encode rationale \(z_{\text{reason}}\).
\item Fuse \([z_{\text{card}},\hat{\mathbf{y}}_{\text{lung}},z_{\text{reason}}]\) to predict \(\hat{y}_{\text{cvd}}\).
\end{enumerate}
\end{tcolorbox}

The agents use deterministic decoding and constrained prompts with fixed output fields to support reproducible text generation. Each output is required to separate (i) observed pulmonary findings, (ii) retrieved mechanism candidates, (iii) synthesized cardiopulmonary effects, and (iv) a bounded cardiovascular rationale. This schema is intentionally narrower than open-ended report generation: the LLM cannot introduce raw image measurements, demographic attributes, treatment history, outcome labels, or a categorical CVD risk prediction that were not supplied by preceding modules. Because the downstream predictor depends on the typed state $\mathcal{S}_{\text{cvd}}$ rather than on a model-specific conversation trace, the same cardiac branch and fusion head can be evaluated with proprietary, open, or local reasoning backends under an identical subject split. The box below summarizes the three fixed contracts used at inference time; only $\Omega_K$ and $\Omega_D$ are LLM prompts, whereas $\Omega_P$ is a controlled perception interface.

\begin{tcolorbox}[
colback=blue!3!white,
colframe=blue!35!black,
colbacktitle=blue!10!white,
coltitle=black,
title={Prompt Contracts for the Reasoning Module},
fonttitle=\bfseries,
boxrule=0.35pt,
arc=1pt,
left=3pt,
right=3pt,
top=3pt,
bottom=3pt]
\footnotesize
\textbf{\(\Omega_P\): Pulmonary Perception Contract.}
\textbf{Input:} LDCT volume \(X\) and a fixed pulmonary concept vocabulary.
\textbf{Output:}
\(\{(c_j,r_j,s_j)\mid s_j>\tau\}\), where each retained finding has a normalized concept, a short descriptor, and a confidence score.
\textbf{Constraint:} pass only structured findings to downstream agents; no raw voxels, identifiers, metadata, or outcome labels are exposed.

\medskip\noindent
\textbf{\(\Omega_K\): Mechanism Discovery Prompt.}
\textbf{Input:} \(R_{\text{lung}}\) and retrieved snippets \(\mathcal{K}_{\text{lung}}\).
\textbf{Instruction:} identify intermediate cardiopulmonary mechanisms supported by the supplied findings and retrieval evidence; do not output a final CVD risk level.
\textbf{Response schema:}
\(\{\texttt{mechanisms}: [(m_1,\text{support}_1),\ldots],
\texttt{summary}: \text{2--3 sentences}\}\).

\medskip\noindent
\textbf{\(\Omega_D\): Diagnostic Synthesis Prompt.}
\textbf{Input:} \(R_{\text{lung}}\), mechanism graph \(G_i\), and \(\mathcal{K}_{\text{lung}}\).
\textbf{Instruction:} synthesize cardiopulmonary effects and a concise rationale; acknowledge uncertainty; do not invent patient history, comorbidities, image measurements, or treatment information.
\textbf{Response schema:}
\(\{\texttt{effects}: [h_1,\ldots],
\texttt{uncertainty}: \text{evidence-limited caveat},
\texttt{rationale}: \text{3--5 sentences}\}\).
\end{tcolorbox}

\subsection{Lung Cancer Risk Prior as Systemic Context}
\label{subsec:lung_cancer_risk_analysis}

While the primary indication for LDCT is lung cancer screening, the malignancy risk inherent in the scan serves as a potent proxy for cumulative biological stress. To leverage this, we integrate a \textbf{Lung Cancer Risk Predictor} as a foundational module, leveraging the established correlation between pulmonary malignancy and cardiovascular vulnerability (e.g., shared pathways of chronic inflammation and oxidative stress).

Given an input volume $X \in \mathbb{R}^{H \times W \times D}$, this module functions as a specialized encoder that distills volumetric texture into a malignancy probability trajectory:
\begin{equation}
    \hat{\mathbf{y}}_{\text{lung}} = \mathcal{F}_{\text{lung}}(X;\theta_{\text{l}}) \in [0,1]^T,
\end{equation}
where $\hat{\mathbf{y}}_{\text{lung}}$ represents a \emph{risk trajectory}: a length-$T$ vector of per-year lung cancer probabilities in $[0,1]$ over a defined temporal horizon (e.g., years 1--6). Unlike manual nodule measurements, this deep learning-based predictor captures subtle, sub-visual parenchymal patterns indicative of long-term disease burden.

Crucially, within our cross-disease framework, $\hat{\mathbf{y}}_{\text{lung}}$ is repurposed not merely as a clinical output, but as a \emph{contextual systemic prior}. By conditioning the downstream cardiovascular model on this validated malignancy signal, we explicitly inject information regarding the patient's global health status and comorbidity profile. This architectural choice reinforces the physiological coherence of the system, allowing the model to modulate its cardiac risk assessment based on the severity of the background pulmonary pathology.

\subsection{Cardiac Localization and Feature Extraction}
\label{subsec:cardiac_extraction}

Directly analyzing the full-field LDCT for cardiovascular risk is suboptimal due to the visual dominance of pulmonary structures and the noise introduced by non-cardiac tissues. To address this, we employ a \textbf{Cardiac Localization and Feature Extraction Module} designed to perform anatomical disentanglement, isolating the heart to enable focused feature learning.

Formally, the module applies a coarse-to-fine localization strategy to extract a region of interest (ROI) centered on the cardiac silhouette:
\begin{equation}
    X_{\text{card}} = \mathcal{D}_{\text{card}}(X;\theta_{\text{d}}) \in \mathbb{R}^{h \times w \times d},
\end{equation}
where the localized subvolume dimensions $(h,w,d)$ are significantly smaller than the original input $(H,W,D)$. This process effectively filters out irrelevant thoracic background (e.g., ribs, spine, and lung parenchyma), ensuring that subsequent convolutional operations are dedicated solely to relevant cardiac morphology.

This targeted extraction is pivotal for two reasons. First, it significantly enhances the signal-to-noise ratio, allowing the encoder to capture fine-grained biomarkers such as \emph{coronary artery calcification (CAC) burden}, \emph{chamber geometry}, and \emph{epicardial adipose tissue volume}. Second, it decouples cardiac feature extraction from pulmonary texture, preventing the model from learning spurious correlations from the lung field. The resulting $X_{\text{card}}$ serves as the anatomical anchor for our multimodal fusion, providing a clean structural representation that complements the semantic reasoning derived from the pulmonary agents.

\subsection{Multimodal Risk Prediction and Optimization}
\label{subsec:fusion_objective}

The final predictor treats visual evidence and reasoning evidence as complementary but separable inputs. For each subject, the lung branch provides the frozen risk trajectory $z_i^{\mathrm{lung}}=\hat{\mathbf{y}}_{\mathrm{lung},i}$, the cardiac branch provides $z_i^{\mathrm{card}}=\mathcal{E}_{\mathrm{card}}(X_{i,\mathrm{card}};\theta_c)$, and the reasoning branch provides $z_i^{\mathrm{reason}}=\mathcal{E}_{\mathrm{text}}(E_{\mathrm{cvd},i})$. We concatenate layer-normalized representations and predict each cardiovascular endpoint with a task-specific head:
\[
\begin{aligned}
h_i &=
\operatorname{LN}(z_i^{\mathrm{card}})
\oplus \operatorname{LN}(z_i^{\mathrm{lung}})
\oplus \operatorname{LN}(z_i^{\mathrm{reason}}),\\
\hat{y}_i^{t} &= \sigma(g_t(h_i;\theta_f)),\quad
t\in\{\mathrm{scr},\mathrm{mort}\}.
\end{aligned}
\]
Here, $\sigma(\cdot)$ is the sigmoid function. The CVD-supervised parameters are restricted to the cardiac encoder and fusion heads, while the lung risk model, pulmonary finding extractor, reasoning interface, and text encoder are frozen or externally instantiated. This separation is important for ablation: replacing $z_i^{\mathrm{reason}}$ with visual lung embeddings, structured findings, fixed rules, or alternative reasoning backends leaves the CVD labels, subject split, and trainable fusion protocol unchanged.

For each endpoint, we optimize the binary cross-entropy objective
\[
\mathcal{L}_{t} =
-\frac{1}{N}\sum_{i=1}^{N}
\left[
y_i^{t}\log \hat{y}_i^{t}
+ (1-y_i^{t})\log(1-\hat{y}_i^{t})
\right].
\]
This objective keeps the generated rationale as an intermediate representation rather than a supervision target: the reasoning branch can guide the decision boundary, but final risk estimation remains anchored by outcome labels and localized cardiac visual evidence.

\section{Experiments}
\label{sec:experiments}

\subsection{Experimental Setup}

\paragraph{Dataset and Preprocessing}
All experiments were conducted on the low-dose CT (LDCT) arm of the \textbf{National Lung Screening Trial (NLST)} dataset~\cite{national2011reduced}, a large randomized study of lung cancer screening in high-risk populations.
After quality control, the final cohort comprised 32{,}136 LDCT scans from 10{,}395 subjects, with each participant undergoing one to three annual scans acquired using heterogeneous scanners and reconstruction kernels.
Following prior work~\cite{chao2021deep}, we considered two cardiovascular prediction tasks:
(1) \textbf{CVD screening}, identifying subjects who developed cardiovascular events during follow-up, and
(2) \textbf{CVD mortality}, identifying deaths attributed to cardiovascular causes based on ICD codes.
Across the full cohort, the screening task included 6{,}339 positive and 25{,}797 negative cases, while the mortality task included 1{,}090 cardiovascular deaths and 31{,}046 non-CVD or censored cases.
Each annual LDCT is evaluated as a scan-level prediction instance, and all scans from the same subject are assigned to the same predefined training, validation, and test split to prevent leakage across screening rounds; the resulting scan counts are reported in Table~\ref{tab:cvd_label_stats}.
All LDCT volumes were standardized for consistent representation across modules, and all reported baselines use subject-level splits to avoid patient leakage.

\paragraph{Comparison Methods}
We compared the proposed framework against representative baselines spanning classical pipelines, deep learning models, and foundation models, including AE+SVM~\cite{chao2021deep}, DeepCAC~\cite{zeleznik2021deep}, KAMP-Net~\cite{guo2019knowledge}, Tri2D-Net~\cite{chao2021deep}, M3FM~\cite{niu2025medical}, and CT-CLIP~\cite{hamamci2024developing}. We include a task-adapted Sybil baseline~\cite{mikhael2023sybil}, implemented as a CVD-supervised prediction head on frozen Sybil representations; in contrast, the \textit{Lung-Risk Only} ablation uses only the frozen Sybil lung-cancer risk score without CVD-supervised adaptation. We also evaluated MedicalNet variants~\cite{chen2019med3d} to assess pulmonary perception, cardiac representation, and cross-disease reasoning.

\paragraph{Implementation}
The lung branch uses frozen Sybil~\cite{mikhael2023sybil} risk scores and CT-CLIP~\cite{hamamci2024developing} pulmonary findings. The cardiac branch localizes ROIs with a pretrained 3D RetinaNet-style detector~\cite{chao2021deep} and encodes cardiac subvolumes with 3D MedicalNet backbones. Reasoning text is generated by deterministic structured prompts and encoded by a frozen ClinicalBERT encoder. All trainable components are optimized with binary cross-entropy using Adam, learning rate $1\times10^{-4}$, batch size 4, early stopping on validation AUC, and identical subject-level splits.

\paragraph{Evaluation Metrics}
Model performance was evaluated using the receiver operating characteristic (ROC) curve and the corresponding area under the curve (AUC), which quantifies overall discriminative ability across decision thresholds. For each task, 95\% confidence intervals (CIs) of the AUC were estimated via non-parametric bootstrapping with 1{,}000 resamples. ROC curves are presented for visual comparison, illustrating sensitivity–specificity trade-offs across competing methods. 

\subsection{Experimental Results}
\label{subsec:main_results}

\begin{table}[!ht]
\centering
\caption{\textbf{Comparison with prior LDCT-based CVD prediction methods.}
We report AUC with 95\% confidence intervals. AE+SVM denotes an autoencoder plus support-vector machine baseline; CAC denotes coronary artery calcium.}
\label{tab:sota_screening_mortality}
\setlength{\tabcolsep}{3.2pt}
\renewcommand{\arraystretch}{1.06}
\footnotesize
\resizebox{0.9\linewidth}{!}{%
\begin{tabular}{lcc}
\toprule
\rowcolor{TableHeader}
\textbf{Method}
& \textbf{CVD Screening AUC}
& \textbf{CVD Mortality AUC} \\
\midrule
AE+SVM~\cite{chao2021deep}
& 0.684 [0.659, 0.711]
& 0.650 [0.605, 0.695] \\
DeepCAC~\cite{zeleznik2021deep}
& 0.753 [0.735, 0.771]
& 0.696 [0.655, 0.737] \\
KAMP-Net~\cite{guo2019knowledge}
& 0.725 [0.700, 0.749]
& 0.671 [0.628, 0.714] \\
CT-CLIP~\cite{hamamci2024developing}
& 0.778 [0.758, 0.796]
& 0.678 [0.629, 0.724] \\
Tri2D-Net~\cite{chao2021deep}
& 0.871 [0.860, 0.882]
& 0.768 [0.734, 0.801] \\
Sybil~\cite{mikhael2023sybil}
& 0.880 [0.861, 0.899]
& 0.794 [0.734, 0.854] \\
M3FM~\cite{niu2025medical}
& 0.892 [0.875, 0.910]
& 0.816 [0.759, 0.874] \\
\midrule
\rowcolor{TableBest}
\textbf{Ours (3D ResNet-50)}
& \textbf{0.919 [0.910, 0.929]}
& \textbf{0.824 [0.797, 0.850]} \\
\rowcolor{TableBest}
\textbf{Ours (3D ResNet-101)}
& \textbf{0.910 [0.889, 0.929]}
& \textbf{0.822 [0.791, 0.850]} \\
\rowcolor{TableBest}
\textbf{Ours (3D ResNet-152)}
& \textbf{0.912 [0.902, 0.922]}
& \textbf{0.833 [0.812, 0.854]} \\
\rowcolor{TableBest}
\textbf{Ours (3D ResNet-200)}
& \textbf{0.916 [0.906, 0.928]}
& \textbf{0.838 [0.815, 0.859]} \\
\bottomrule
\end{tabular}
}
\vspace{-0.5em}
\end{table}

\paragraph{Comparison with state-of-the-art methods}

Table~\ref{tab:sota_screening_mortality} summarizes the performance of our framework against representative state-of-the-art approaches on both \textbf{CVD Screening} and \textbf{CVD Mortality} tasks. Traditional pipelines such as AE+SVM and early CNN-based models (DeepCAC, KAMP-Net) show limited discrimination, with AUCs below 0.76 for screening and 0.70 for mortality, suggesting that handcrafted or purely image-based representations capture only partial cardiovascular information. Tri2D-Net improves performance by aggregating multi-view features, achieving an AUC of 0.871 for screening and 0.768 for mortality, but remains limited by the lack of multimodal contextual reasoning. More advanced models such as Sybil and M3FM, which use large-scale pretraining and vision--language integration, further improve discrimination, reaching AUCs of 0.880--0.892 for screening and 0.794--0.816 for mortality. Our proposed framework surpasses all competing methods across both tasks. With a 3D ResNet-50 backbone, the model achieves an AUC of 0.919 [0.910--0.929] for CVD screening and 0.824 [0.797--0.850] for CVD mortality prediction, outperforming the strongest baseline (M3FM) by absolute gains of +2.7 and +0.8 points, respectively. Performance remains stable across deeper MedicalNet variants (ResNet-101/152/200), indicating robustness to network depth and model capacity. Notably, these gains are achieved without explicit coronary calcium scoring or handcrafted cardiac priors, underscoring the advantage of integrating pulmonary findings, cardiac subvolume encoding, and knowledge-guided indicator reasoning in a unified modular framework.

Overall, these results validate that our framework effectively leverages cross-disease signals embedded in LDCT to achieve state-of-the-art discrimination while preserving physiologically grounded interpretability through explicit pulmonary-to-cardiac reasoning. The comparison is best understood by task structure rather than model size alone. Foundation models such as CT-CLIP and M3FM learn broad representations from radiology text, images, and multiple clinical tasks, but their cardiovascular predictions are still produced through relatively opaque downstream heads. Our framework instead introduces an intermediate knowledge-management layer: pulmonary findings are normalized, linked to retrieved cardiopulmonary mechanisms, converted into bounded rationales, and fused with localized cardiac morphology. This makes the cross-disease pathway explicit enough for ablation, auditing, and swapping across reasoning backends, supporting the premise that LDCT-based CVD assessment benefits from both visual representation learning and structured inter-organ reasoning.

\paragraph{Ablation Study and Module Evaluation}

\begin{table*}[t]
\centering
\caption{\textbf{Ablation study on CVD screening and CVD mortality prediction from LDCT.}
We report AUC with 95\% confidence intervals across MedicalNet backbones (3D ResNet-50/101/152/200).
Rows compare single-region visual branches and progressive fusion with lung-risk and reasoning signals; Lung-Risk denotes the frozen lung-cancer risk prior, and Reasoning denotes the pulmonary-to-cardiac rationale embedding.}
\label{tab:cvd_ablation}
\setlength{\tabcolsep}{6.5pt}
\renewcommand{\arraystretch}{1.15}
\small
\resizebox{0.8\linewidth}{!}{%
\begin{tabular}{lcccc}
\toprule
\rowcolor{TableHeader}
\multicolumn{5}{c}{\textbf{CVD Screening AUC [95\% CI]}} \\
\midrule
\textbf{Variant} & \textbf{3D ResNet-50} & \textbf{3D ResNet-101} & \textbf{3D ResNet-152} & \textbf{3D ResNet-200} \\
\midrule
Lung Region Only
& 0.730 [0.705, 0.754] & 0.702 [0.664, 0.740] & 0.714 [0.687, 0.740] & 0.708 [0.685, 0.729] \\
Cardiac Region Only
& 0.886 [0.871, 0.901] & 0.875 [0.857, 0.892] & 0.881 [0.866, 0.896] & 0.886 [0.871, 0.900] \\
+ Lung-Risk
& 0.899 [0.885, 0.914] & 0.889 [0.872, 0.905] & 0.894 [0.880, 0.908] & 0.902 [0.888, 0.916] \\
\rowcolor{TableBest}
\textbf{+ Lung-Risk \& Reasoning}
& \textbf{0.919 [0.910, 0.929]} & \textbf{0.910 [0.889, 0.929]} & \textbf{0.912 [0.902, 0.922]} & \textbf{0.916 [0.906, 0.928]} \\
\bottomrule
\end{tabular}
}
\vspace{0.8em}
\resizebox{0.8\linewidth}{!}{%
\begin{tabular}{lcccc}
\rowcolor{TableHeader}
\multicolumn{5}{c}{\textbf{CVD Mortality AUC [95\% CI]}} \\
\midrule
\textbf{Variant} & \textbf{3D ResNet-50} & \textbf{3D ResNet-101} & \textbf{3D ResNet-152} & \textbf{3D ResNet-200} \\
\midrule
Lung Region Only
& 0.658 [0.620, 0.696] & 0.657 [0.617, 0.694] & 0.677 [0.636, 0.718] & 0.660 [0.620, 0.698] \\
Cardiac Region Only
& 0.772 [0.740, 0.804] & 0.776 [0.743, 0.808] & 0.782 [0.747, 0.817] & 0.769 [0.734, 0.803] \\
+ Lung-Risk
& 0.802 [0.771, 0.830] & 0.801 [0.769, 0.834] & 0.797 [0.766, 0.828] & 0.803 [0.773, 0.833] \\
\rowcolor{TableBest}
\textbf{+ Lung-Risk \& Reasoning}
& \textbf{0.824 [0.797, 0.850]} & \textbf{0.822 [0.791, 0.850]} & \textbf{0.833 [0.812, 0.854]} & \textbf{0.838 [0.815, 0.859]} \\
\bottomrule
\end{tabular}
}
\end{table*}

\begin{figure}[!ht]
\centerline{\includegraphics[width=\columnwidth]{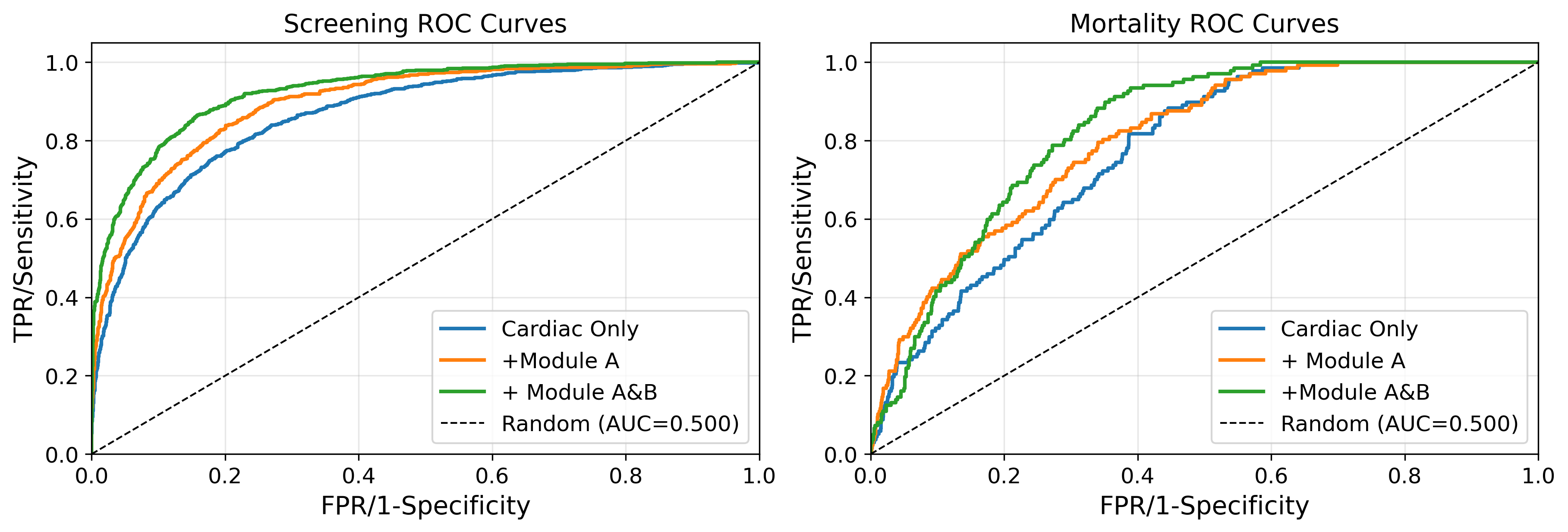}}
\caption{\mdseries
\textbf{Receiver operating characteristic (ROC) curves for CVD screening (left) and CVD mortality prediction (right).}
 We compare progressively enhanced variants. In the plot legend, \textit{Module A} denotes the frozen lung-risk prior, and \textit{Module B} denotes the pulmonary-to-cardiac reasoning branch; thus \textit{+ Module A\&B} corresponds to \textit{+ Lung-Risk \& Reasoning}.}
\label{fig:roc_curves}
\end{figure}

Table~\ref{tab:cvd_ablation} and Fig.~\ref{fig:roc_curves} quantify the incremental contribution of each component across MedicalNet backbones.
Consistent with the ablation results, \textit{Lung Region Only} provides limited discrimination (screening AUC 0.70--0.73; mortality AUC 0.65--0.68), indicating that pulmonary visual evidence alone carries insufficient cardiovascular signal.
Replacing this with \textit{Cardiac Region Only} markedly improves both endpoints (average +0.17 AUC for screening and +0.12 for mortality), highlighting the value of localized 3D cardiac morphology such as coronary calcification and pericardial fat. Augmenting the cardiac branch with pulmonary context (\textit{+ Lung-Risk}) yields further gains (e.g., 0.886 $\rightarrow$ 0.899 screening AUC with ResNet-50), demonstrating that pulmonary comorbidity information complements cardiac structure. Fig.~\ref{fig:roc_curves} further shows that these improvements persist across a wide range of operating points; the legacy plot labels \textit{Module A} and \textit{Module B} correspond to the lung-risk prior and reasoning branch, respectively. The \textit{+ Lung-Risk \& Reasoning} variant consistently exhibits the most left-shifted and steepest ROC curves for both screening and mortality, indicating superior sensitivity–specificity trade-offs beyond what can be achieved by image features or lung-risk alone.

\begin{figure}[!ht]
\centerline{\includegraphics[width=\columnwidth]{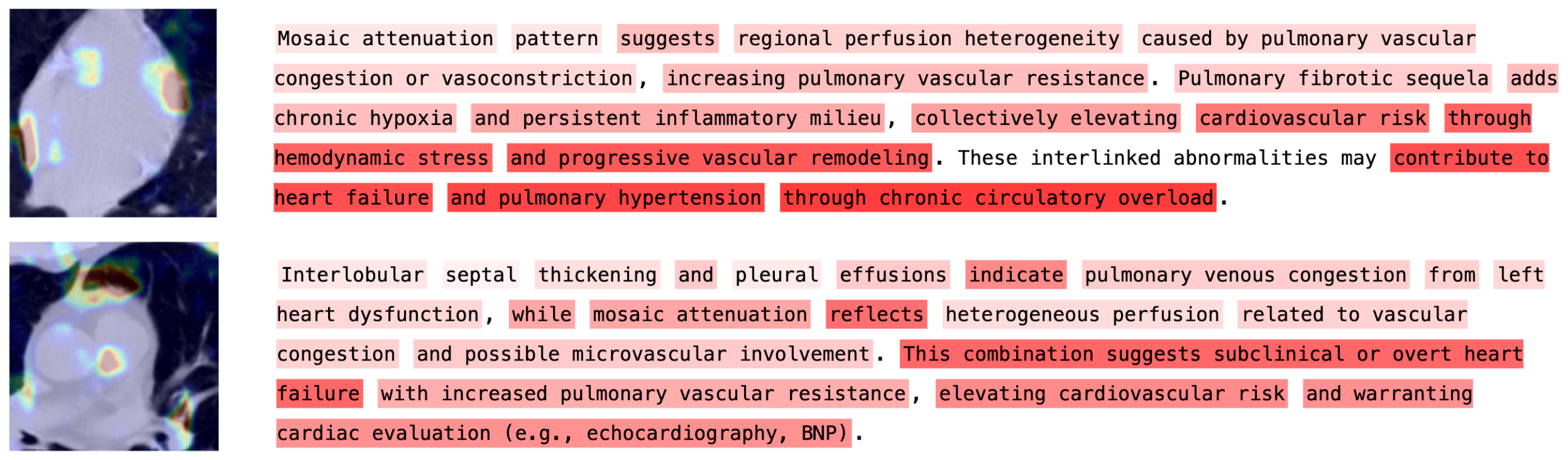}}
\caption{\mdseries \textbf{Visualization of explainable cross-disease reasoning.}
 \textbf{Left:} Grad-CAM activation over the 3D cardiac subvolume, highlighting regions most influential for CVD prediction. \textbf{Right:} token-level attribution map from the pulmonary-to-cardiac reasoning pathway, indicating salient phrases contributing to the decision.}
\label{fig:explanation_vis}
\end{figure}

\paragraph{Visualization of Explainable Cross-Disease Reasoning}
\label{subsubsec:visual_explanation}

To examine how the proposed framework integrates multimodal evidence, we visualize the learned reasoning process using 3D gradient-weighted class activation mapping (Grad-CAM~\cite{selvaraju2017grad}) for the \textit{3D cardiac encoder} and token-level attribution for the \textit{text-based reasoning pathway}. As shown in Fig.~\ref{fig:explanation_vis}, Grad-CAM highlights anatomically meaningful cardiac subregions, including ventricular walls, pericardial boundaries, and coronary arteries, indicating that the cardiac branch focuses on localized morphological biomarkers relevant to cardiovascular impairment. 
In parallel, the textual attribution map assigns high importance to clinically meaningful phrases such as \textit{vascular remodeling}, \textit{pulmonary congestion}, and \textit{hemodynamic overload}, reflecting system-level cardiopulmonary mechanisms inferred from pulmonary findings. Taken together, these visualizations suggest that model predictions use complementary evidence sources: localized structural cues captured by the cardiac encoder and mechanistic, physiology-driven signals encoded by the reasoning pathway. This multimodal consistency provides transparent insight into how pulmonary abnormalities may inform cardiovascular risk estimation from LDCT, while the control studies in \S\ref{subsubsec:reasoning_effect} are used to test whether the performance gain exceeds vision-only, semantic-only, fixed-rule, and backend-specific alternatives.

\paragraph{Component-Wise Evaluation and Full-Model Performance}
\label{subsubsec:component_only_eval}

\begin{table}[!ht]
\centering
\caption{\textbf{Performance of individual components and their integration} on CVD screening and CVD mortality prediction.
The cardiac branch uses the 3D ResNet-50 backbone when applicable; Lung-Risk is the frozen lung-cancer risk prior, and Reasoning is the encoded pulmonary-to-cardiac rationale.}
\label{tab:component_only}

\setlength{\tabcolsep}{7pt}
\renewcommand{\arraystretch}{1.15}
\small

\begin{adjustbox}{max width=\linewidth}
\begin{tabular}{lcc}
\toprule
\rowcolor{TableHeader}
\textbf{Model Component}
& \textbf{CVD Screening AUC [95\% CI]}
& \textbf{CVD Mortality AUC [95\% CI]} \\
\midrule
Lung-Risk Only
& 0.713 [0.667, 0.759]
& 0.670 [0.647, 0.692] \\

Reasoning Only
& 0.832 [0.814, 0.850]
& 0.776 [0.752, 0.799] \\

Cardiac Region Only
& 0.886 [0.871, 0.901]
& 0.772 [0.740, 0.804] \\

\rowcolor{TableBest}
\textbf{Full Model (Ours)}
& \textbf{0.919 [0.910, 0.929]}
& \textbf{0.824 [0.797, 0.850]} \\
\bottomrule
\end{tabular}
\end{adjustbox}
\end{table}

Table~\ref{tab:component_only} compares individual components and their combination. \textit{Lung-Risk Only} provides moderate discrimination (AUC 0.713/0.670), reflecting shared exposure and inflammatory context. \textit{Reasoning Only} improves to 0.832/0.776 because it encodes CT-CLIP-derived pulmonary findings after knowledge-grounded mechanism synthesis, rather than text detached from the scan. The \textit{Cardiac Region Only} branch reaches 0.886/0.772 by capturing localized anatomical biomarkers. Combining all three components yields 0.919/0.824, showing that pulmonary risk priors, reasoning-derived physiological context, and cardiac morphology provide complementary evidence.

\paragraph{Effect of Pulmonary-to-Cardiac Reasoning and Control Variants}
\label{subsubsec:reasoning_effect}

\begin{figure}[!ht]
\centering
\includegraphics[width=\columnwidth]{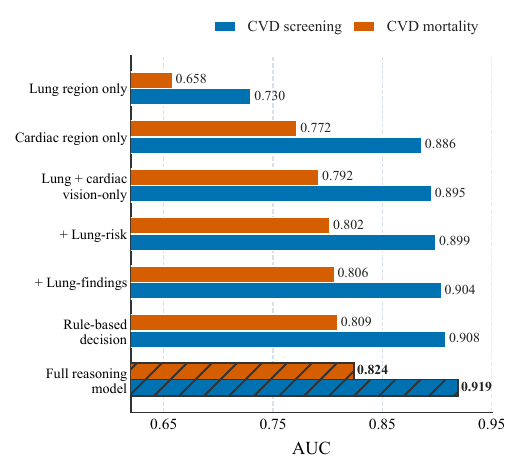}
\caption{\textbf{Control study isolating cross-disease reasoning} using the 3D ResNet-50 backbone. Variants progress from single-branch and vision-only inputs to structured pulmonary findings, fixed expert rules, and the full reasoning model. Hatched bars denote the full model.}
\label{fig:reasoning_effect}
\end{figure}

Fig.~\ref{fig:reasoning_effect} compares ways of integrating pulmonary information. Vision-only fusion improves only modestly over the cardiac branch (0.886 $\rightarrow$ 0.895 screening AUC), so the gain is not explained by a larger thoracic field. Structured pulmonary semantics and fixed rules are stronger but still intermediate, reaching 0.904/0.806 and 0.908/0.809, respectively. The full reasoning model reaches 0.919/0.824 by converting patient-specific findings into knowledge-grounded mechanisms before fusion. We interpret this gain as complementary to fixed rules rather than as a replacement for them: fixed rules test a small set of expert-specified associations, while the reasoning branch can represent borderline, co-occurring, or heterogeneous finding combinations under the same downstream fusion protocol.

\paragraph{Robustness to Pulmonary Finding Perturbations}
\label{subsubsec:finding_robustness}

\begin{figure}[!ht]
\centering
\includegraphics[width=\columnwidth]{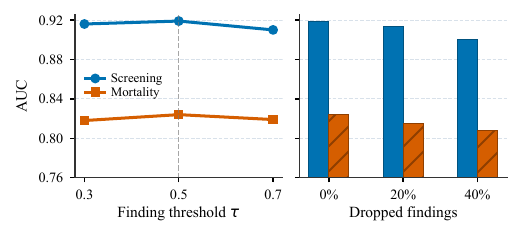}
\caption{\textbf{Robustness to pulmonary finding perturbations.} We report absolute AUCs under threshold changes and random finding dropout. The default threshold $\tau=0.5$ and no-dropout setting achieve the strongest performance, while perturbations lead to gradual degradation.}
\label{fig:finding_robustness}
\end{figure}

Fig.~\ref{fig:finding_robustness} tests sensitivity to pulmonary finding extraction. Loosening the CT-CLIP threshold to $\tau=0.3$ introduces additional low-confidence findings and yields 0.916/0.818, while tightening it to $\tau=0.7$ reduces finding coverage and yields 0.910/0.819. Dropping 20\% and 40\% of findings causes gradual degradation (0.913/0.815 and 0.900/0.808), suggesting that the model remains competitive under moderate upstream incompleteness rather than collapsing under modest perception noise. These perturbations quantify error propagation through the structured finding interface, complementing future radiologist-level validation.

\paragraph{Reasoning Backend Robustness and Deployment}
\label{subsubsec:backend_robustness}

\begin{table}[!ht]
\centering
\caption{\textbf{Reasoning backend ablation} on the 3D ResNet-50 backbone. All values are AUC with 95\% confidence intervals.}
\label{tab:reasoning_agent_ablation}
\setlength{\tabcolsep}{5.8pt}
\renewcommand{\arraystretch}{1.08}
\small
\begin{adjustbox}{max width=\linewidth}
\begin{tabular}{lcc}
\toprule
\rowcolor{TableHeader}
\textbf{Backend}
& \textbf{CVD Screening AUC [95\% CI]}
& \textbf{CVD Mortality AUC [95\% CI]} \\
\midrule
GPT-5-nano
& \textbf{0.919 [0.910, 0.929]}
& 0.824 [0.797, 0.850] \\
Qwen-Flash
& 0.912 [0.900, 0.923]
& 0.818 [0.792, 0.844] \\
Baichuan-M2
& 0.918 [0.911, 0.925]
& \textbf{0.827 [0.805, 0.849]} \\
\bottomrule
\end{tabular}
\end{adjustbox}
\end{table}

Table~\ref{tab:reasoning_agent_ablation} evaluates dependence on the reasoning backend. GPT-5-nano is the default, Qwen-Flash tests an alternative Qwen-family backend, and Baichuan-M2 tests a locally deployable medical LLM. All preserve the same trend, with Baichuan-M2 nearly matching GPT-5-nano on screening and slightly improving mortality. This supports two practical conclusions. First, the gain is driven primarily by the structured pulmonary-to-cardiac formulation rather than by a single proprietary prior. Second, privacy-sensitive deployments can instantiate the reasoning stage locally without sending raw CT volumes, patient identifiers, or clinical metadata to external services. Because the reasoning input is a compact finding set rather than a full 3D image, the generated state can also be cached and reused for repeated endpoint prediction.

\paragraph{Cohort Statistics and Reproducibility Details}
\label{subsubsec:reproducibility_details}

\begin{table}[!ht]
\centering
\caption{\textbf{Scan-level label distribution} for the two cardiovascular prediction tasks after quality control, under subject-level train/validation/test splits.}
\label{tab:cvd_label_stats}
\setlength{\tabcolsep}{4.8pt}
\renewcommand{\arraystretch}{1.04}
\footnotesize
\begin{tabular}{lcccc}
\toprule
\rowcolor{TableHeader}
& \multicolumn{2}{c}{CVD Screening} & \multicolumn{2}{c}{CVD Mortality} \\
\cmidrule(lr){2-3}\cmidrule(lr){4-5}
\textbf{Split} & Positive & Negative & Positive & Negative \\
\midrule
Train & 5{,}492 & 23{,}643 & 953 & 28{,}182 \\
Validation & 284     & 715     & 55  & 944 \\
Test  & 563     & 1{,}439  & 82  & 1{,}920 \\
\bottomrule
\end{tabular}
\end{table}

All experiments use the same subject-level train/validation/test split to prevent patient leakage across annual LDCT scans. Pulmonary risk scores come from frozen Sybil, cardiac regions from a pretrained 3D RetinaNet-style detector, pulmonary findings from CT-CLIP, and reasoning text from ClinicalBERT. Trainable components use binary cross-entropy with Adam, validation-AUC early stopping, and 1{,}000 subject-level bootstrap resamples for confidence intervals. During inference, pulmonary findings and reasoning outputs can be cached, so repeated cardiovascular assessment only requires lightweight fusion once intermediate representations have been generated. The fixed stage interfaces also allow alternative pulmonary descriptors, rules, or reasoning backends to be tested without changing splits or labels.

\begin{table}[!ht]
\centering
\caption{\textbf{Joint pulmonary--CVD outcome statistics} on held-out test scans. Counts are reported for the subject-level test split; Lung finding $+/-$ denotes presence/absence of a pulmonary abnormality indicator.}
\label{tab:lung_cvd_joint}
\setlength{\tabcolsep}{5.2pt}
\renewcommand{\arraystretch}{1.03}
\footnotesize
\begin{tabular}{lcccc}
\toprule
\rowcolor{TableHeader}
& \multicolumn{2}{c}{CVD Screening} & \multicolumn{2}{c}{CVD Mortality} \\
\cmidrule(lr){2-3}\cmidrule(lr){4-5}
\textbf{Group} & Positive & Negative & Positive & Negative \\
\midrule
Lung finding + & 33 & 74 & 8 & 99 \\
Lung finding - & 530 & 1{,}365 & 74 & 1{,}821 \\
\bottomrule
\end{tabular}
\end{table}

Table~\ref{tab:lung_cvd_joint} provides a cohort-level sanity check on held-out subjects. The counts are not used as causal evidence; they simply show co-occurrence patterns that motivate treating pulmonary findings as structured context.

\paragraph{Efficiency Considerations}
\label{subsubsec:efficiency}

\begin{table}[!ht]
\centering
\caption{\textbf{Amortized inference characteristics} for visual encoding, reasoning, and fusion. LLM denotes large language model.}
\label{tab:quant_efficiency}
\setlength{\tabcolsep}{4.6pt}
\renewcommand{\arraystretch}{1.02}
\footnotesize
\resizebox{\linewidth}{!}{%
\begin{tabular}{lcc}
\toprule
\rowcolor{TableHeader}
\textbf{Component} & \textbf{Latency} & \textbf{Role} \\
\midrule
M3FM end-to-end & $\sim$120 ms & Full-volume 3D forward \\
Sybil lung-risk prior & $\sim$90 ms & Frozen pulmonary risk prior \\
MedicalNet cardiac encoder & $\sim$8 ms & Local cardiac ROI encoding \\
CT-CLIP pulmonary encoder & $\sim$220 ms & Pulmonary perception \\
Fusion predictor & $\sim$2--5 ms & Final classification \\
Reasoning LLM & $\sim$1.5--3.5 s & Uncached mechanism generation \\
\bottomrule
\end{tabular}
}
\end{table}

Visual timings in Table~\ref{tab:quant_efficiency} are measured on a single NVIDIA H100 80GB GPU; reasoning latency is measured after pulmonary findings are extracted. The LLM stage adds uncached latency, but it operates on compact text rather than raw 3D volumes and can be cached for repeated endpoint queries. Combined with the backend ablation in Table~\ref{tab:reasoning_agent_ablation}, this also shows a practical deployment path: local medical LLMs such as Baichuan-M2 preserve near-identical AUC while avoiding external API round trips, making latency more controllable and often lower depending on hardware. Thus, the framework is intended for retrospective screening workflows where interpretability and auditability are valuable, rather than for sub-second real-time triage.

\paragraph{Rationale Safeguards and Audit}
\label{subsubsec:rationale_safeguards}

The reasoning prompts are constrained to reduce unsupported patient-specific claims. The Knowledge Reasoning Agent receives only structured pulmonary findings and returns intermediate mechanisms; the Diagnostic Reasoning Agent then produces a concise rationale conditioned on those mechanisms. Both prompts require clinical plausibility, uncertainty acknowledgment, and avoidance of invented patient history. We audit rationales at three levels: finding support by $R_{\text{lung}}$, mechanism support by retrieved cardiopulmonary knowledge, and prediction consistency with the cardiac branch and final fused score. This does not replace blinded clinician review, but it makes unsupported findings, generic mechanisms, and prediction-inconsistent rationales visible.

\paragraph{Clinical Workflow Interpretation}
\label{subsubsec:workflow_interpretation}

The workflow uses an LDCT scan already acquired for lung screening, without requiring a second cardiac CT or manual calcium scoring. Its modular outputs separate localized cardiac evidence, systemic pulmonary context, and mechanism-level reasoning. This separation makes evidence sources inspectable and allows improved finding extractors, local retrieval corpora, or institution-specific reasoning backends to be substituted under the same downstream evaluation protocol. Importantly, the reasoning branch is not used as an unconstrained clinical oracle: it is a structured intermediate representation that must remain consistent with the extracted pulmonary findings and the final fused prediction. This design preserves the practical advantage of automated risk assessment while making the cross-organ evidence pathway easier to review, debug, and update as stronger perception models or local medical language models become available.

\section{Limitations}
\label{sec:limitations}

This study is evaluated on NLST, so external LDCT cohorts are needed to assess generalization across scanners, populations, and clinical settings. The generated rationales are mechanism-level explanations rather than causal proof or standalone clinical evidence, and structured pulmonary findings may omit finer quantitative descriptors such as emphysema extent or spatial distribution. Beyond external generalization, several issues remain open. First, although structured prompts constrain agent outputs, the generated rationales are post-hoc explanations whose faithfulness to the classifier's actual decision path is not guaranteed and should be examined against feature-level interventions. Second, AUC characterizes ranking quality but not calibration, so prospective deployment will require local recalibration and cost-sensitive threshold selection. Third, our cohort is restricted to NLST-eligible high-risk smokers, leaving subgroup behavior across age, sex, and pack-year strata to be characterized in future work. Overall, finding extraction, calibration, subgroup performance, rationale faithfulness, and clinician-rated rationale usefulness all warrant prospective multi-center validation.

\section{Conclusion}
\label{sec:conclusion_and_discussion}

We proposed an \emph{Explainable Cross-Disease Reasoning Framework} for integrated pulmonary and cardiovascular risk assessment from low-dose chest CT. By combining a lung-risk prior, pulmonary perception, pulmonary-to-cardiac reasoning, localized cardiac feature encoding, and multimodal fusion, the framework models lung--heart interactions while exposing interpretable intermediate evidence. Experiments on NLST show consistent gains over cardiac-only, vision-only, rule-based, and foundation-model baselines, while targeted controls indicate that the improvement is not simply due to extra thoracic image area, fixed rule propagation, or a single reasoning backend. The modular design separates risk prior, perception, reasoning, and cardiac evidence, making each source auditable and replaceable as stronger local models become available. More broadly, this template extends to other organ-pair cross-disease prediction problems beyond cardiovascular screening.

\bibliographystyle{ACM-Reference-Format}
\bibliography{main}

\end{document}